\documentclass[conference]{IEEEtran}

\usepackage{cite}
\usepackage{amsmath,amssymb,amsfonts}
\usepackage{algorithmic}
\usepackage{algorithm}
\usepackage{graphicx}
\usepackage{textcomp}
\usepackage{xcolor}
\usepackage{url}

\usepackage{booktabs}
\usepackage{tabularx}
\usepackage{multirow}
\usepackage{adjustbox}
\usepackage{array}
\usepackage{hhline}

\newcolumntype{C}[1]{>{\arraybackslash}m{#1}}

\begin{document}
\title{Empowering Prior to Court Legal Analysis: A Transparent and Accessible Dataset for Defensive Statement Classification and Interpretation}
\date{}


\author{
\IEEEauthorblockN{Yannis Spyridis, Jean-Paul Younes}
\IEEEauthorblockA{Kingston University London \\ London, United Kingdom \\ \{y.spyridis, j.younes\}@kingston.ac.uk\\}
\and
\IEEEauthorblockN{Haneen Deeb}
\IEEEauthorblockA{University of Portsmouth \\ Portsmouth, United Kingdom \\ haneen.deeb@port.ac.uk\\}
\and
\IEEEauthorblockN{Vasileios Argyriou}
\IEEEauthorblockA{Kingston University London \\ London, United Kingdom \\ vasileios.argyriou@kingston.ac.uk\\}
}

\maketitle

\begin{abstract}
    The classification of statements provided by individuals during police interviews is a complex and significant task within the domain of natural language processing (NLP) and legal informatics. The lack of extensive domain-specific datasets raises challenges to the advancement of NLP methods in the field. This paper aims to address some of the present challenges by introducing a novel dataset tailored for classification of statements made during police interviews, prior to court proceedings. Utilising the curated dataset for training and evaluation, we introduce a fine-tuned DistilBERT model that achieves state-of-the-art performance in distinguishing truthful from deceptive statements. To enhance interpretability, we employ explainable artificial intelligence (XAI) methods to offer explainability through saliency maps, that interpret the model's decision-making process. Lastly, we present an XAI interface that empowers both legal professionals and non-specialists to interact with and benefit from our system. Our model achieves an accuracy of 86\%, and is shown to outperform a custom transformer architecture in a comparative study. This holistic approach advances the accessibility, transparency, and effectiveness of statement analysis, with promising implications for both legal practice and research.
\end{abstract}

\section{Introduction}
The legal system is built upon the principles of justice and the rule of law, and the settlement of legal disputes often depends on the strength of the presented arguments. Individuals facing accusations typically put forth statements and arguments to either defend themselves or mitigate their liability. The classification of statements made in this context involves their automated identification and categorisation into distinct classes. It plays a vital role in ensuring a fair process by enabling legal professionals to better understand and evaluate their case. This classification is essential for legal practitioners to navigate the complexities of criminal cases, enabling them to distinguish between various argument strategies, assess their validity, and ultimately facilitate the pursuit of justice.

Classifying statements made by accused individuals presents a series of distinctive challenges. Firstly, the language used in these statements may vary widely, from straightforward admissions or denials to subtle explanations, justifications, or even appeals to sympathy. The diversity in language and argumentation strategies necessitates a refined approach to classification, which is often complicated by the emotional and psychological factors at play in criminal cases \cite{salerno2009emotional}. Furthermore, the need to protect the rights and dignity of the accused imposes an ethical dimension on this classification task, requiring sensitivity and respect for privacy. The task fundamentally involves distinguishing between credible statements and false or misleading claims, and presents an inherent difficulty. In addition, the need for interpretability and transparency in legal decision-making \cite{deeks2019judicial}, calls for methods that not only classify arguments but also provide insights into the reasoning behind the classification, which adds an additional layer of complexity to the task \cite{vilone2020explainable}.

Various methodologies can be employed for the automated classification of statements made by accused individuals. These may include rule-based systems, supervised machine learning models, or sentiment analysis \cite{lawrence2020argument}. While human professionals also engage in this task, their involvement can introduce subjective biases and inconsistencies into the classification process \cite{thompson2017interpretation}. The applications of this process extend to legal research, case assessment, and criminal justice system improvement. Additionally, the incorporation of legal domain knowledge and contextual information is often important to enhance the accuracy of classification. Accurate categorisation of arguments can aid legal professionals in evaluating the credibility and relevance of statements made by accused individuals, assisting in plea bargaining, and ensuring that due process is adhered to \cite{re2019developing}. Furthermore, it can contribute to a more efficient and fair criminal justice system by streamlining the review of statements and enhancing the understanding of the defendant's perspectives. Inaccurate assessments of statements can have profound legal and societal implications, potentially leading to unjust legal outcomes, misallocated resources, and undermined public trust in the legal system. Therefore, the need for robust, interpretable, and accessible tools in this domain is significant \cite{cui2023survey}.

This paper represents a novel effort in addressing the critical issue of classifying statements made during police interviews through a holistic approach that leverages state-of-the-art natural language processing (NLP) techniques and user-centric design principles. At its core, our primary objective is to significantly enhance the accuracy and transparency of statement analysis, thereby equipping legal professionals, researchers, and stakeholders with a powerful tool for making more informed and equitable legal decisions. To achieve this goal, we introduce a novel dataset of transcripts from police interviews, which provides a valuable resource for the advancement of NLP research in the legal domain. Additionally, we develop a fine-tuned DistilBERT model tailored specifically for the classification of these arguments, ensuring the highest levels of accuracy in categorisation. Notably, our research goes beyond mere classification, as we also incorporate explainable artificial intelligence (XAI) techniques to shed light on the model's decision-making process, promoting transparency and trust in the system. Lastly, in an effort to make our research accessible and practical, we develop a user-friendly XAI interface, bridging the gap between sophisticated NLP technology and end-users in the legal field. Through this holistic approach, our research aims to contribute to the advancement of NLP and legal informatics, while fostering a more just and efficient legal system.

In summary, the key contributions of this paper are the following:

\begin{itemize}
    \item It introduces a novel and carefully curated dataset of statements made during police interviews. This dataset serves as the foundation for the research and enables the development and evaluation of the classification model.

    \item It develops a fine-tuned DistilBERT model, refined to the task of statement classification. The model achieves high performance in distinguishing truthful from deceptive statements.

    \item It employs state-of-the-art explainability visualisation techniques to enhance the interpretability of the model's decisions. These visualisations provide insight into the rationale behind the model's predictions, allowing users to understand and trust the model's outputs.

    \item It presents an advanced user-friendly interface to cover the need for XAI in both legal practice and research. This interface enables users of any background to interact with the classification system, making statement analysis more accessible and intuitive.
\end{itemize}

The remainder of this paper is structured as follows: Section \ref{sec:related_work} provides NLP background and reviews relevant research conducted in the legal domain on statement classification. Section \ref{sec:methodology} presents the methodology that was followed. including background on the dataset and the explainability methods. Section \ref{sec:evaluation} discusses the results of this work and presents the developed XAI interface. Section \ref{sec:conclusion} provides the conclusion of the paper.

\section{Related Work}
\label{sec:related_work}

Like in many areas, NLP has emerged as a strong tool within the legal domain, revolutionising the way professionals and researchers interact with text-based statements and legal documents \cite{zhong2020does}. NLP methods enable the automated analysis of legal texts, including contracts, court cases, or statements prior to court proceedings. These techniques facilitate tasks such as information extraction, fake news detection \cite{shushkevich2021detecting}, argument classification, and legal document summarisation \cite{merchant2018nlp}. Therefore, NLP can significantly accelerate the research process and enhance the efficiency of legal practitioners by providing tools for information retrieval, knowledge discovery, and evidence assessment \cite{re2019developing}. Furthermore, NLP-powered applications are instrumental in ensuring legal compliance, contract management, and the automation of routine legal tasks, reshaping the landscape of the legal field in novel ways.

\subsection{Transformer architectures}
The transformer model \cite{vaswani2017attention} is a deep learning architecture that relies on self-attention mechanisms to capture contextual relationships between words in a text. This architecture has achieved state-of-the-art performance on a wide range of NLP tasks, including text classification in the legal domain. One of the key advantages of the transformer architecture is its ability to capture long-range dependencies in text, which is particularly important in legal arguments. The transformer's self-attention mechanism allows it to attend to different parts of the input text and weigh the importance of each word in the context of the entire text \cite{devlin2018bert}. This enables the model to effectively capture the semantic meaning and context of statements, leading to improved classification accuracy \cite{zheng2021rethinking}. 

In the context of binary classification, transformers can be used to classify input data into one of two classes. The input data is typically represented as a sequence of tokens, such as words or characters, and each token is embedded into a continuous vector representation. The transformer model then processes these embeddings through a series of self-attention layers and feed-forward layers, allowing it to capture the relevant information for the classification task \cite{abnar2020quantifying}, \cite{tay2021synthesizer}. During training, the transformer model learns to optimise its parameters by minimising a loss function, such as binary cross-entropy, which measures the discrepancy between the predicted probabilities and the true labels. The model updates its parameters using backpropagation and gradient descent, iteratively improving its performance on the training data \cite{wolf2020transformers}. The key advantage of transformers in this context is their ability to handle variable-length input sequences. The self-attention mechanism allows the model to attend to different parts of the input sequence, regardless of their position, enabling it to capture long-range dependencies and contextual information \cite{tay2020long}.

\subsection{Classification in the legal domain}

Several studies have investigated AI-assisted classification in the legal domain across various subjects. The study in \cite{song2022multi} addresses challenges in legal multi-label document classification by introducing a new dataset of legal opinions with manually labeled legal procedural postures. A domain-specific pre-trained deep learning architecture is presented, with a label-attention mechanism that showcases efficacy in overcoming data scarcity and class imbalance issues.

Court case transcripts are analysed in \cite{ratnayaka2019classifying} based on discourse and argumentative properties. The study investigates the utilisation of discourse relationships and sentence properties to extract relevant information from the transcripts, addressing the significance of such data for the legal domain. It proposes a classification framework that employs a combination of machine learning models and rule-based approaches to categorise sentence pairs based on their observed relationship types and distinguish them as contributing to legal arguments or not.

The task of legal judgment prediction, which involves automatically predicting the outcome of a court case based on the text describing the case's facts, is investigated in \cite{chalkidis2019neural}. The study introduces a new legal judgment prediction dataset, comprising cases from the European Court of Human Rights. An evaluation of several neural models is conducted on this dataset, leading to the establishment of robust performance benchmarks that outperform previous feature-based models across binary and multi-label classification and case importance prediction.

\subsection{Model interpretability in NLP}

Despite the success of state-of-the-art language models, interpretability remains a significant challenge within this area of research. Language models are often considered "black boxes" because it is arduous to discern the internal mechanisms and feature representations that lead to specific predictions. Some approaches focus on local explanations, which aim to identify the most important features or words that contribute to a model's prediction. For instance, input saliency methods have been used to explain predictions of deep learning models in NLP \cite{ronnqvist2022explaining}. These methods highlight the words or features that have the most influence on the model's output.

Moreover, various studies have delved into the influence of model architecture and design decisions on interpretability. For instance, approaches that focus explicit word interaction graph layers have been proposed to enhance interpretability by capturing intricate word interactions \cite{sekhon2023improving}. Furthermore, studies have explored the incorporation of likelihood considerations in NLP classification explanations, aiming to offer more insightful and easily understandable explanations \cite{harbecke2020considering}.

\section{Methodology}
\label{sec:methodology}

\subsection{Produced Dataset}

\begin{table}[b]
    \small
    \caption {Dataset preview. }
    \label{tab:dataset}
    {\renewcommand{\arraystretch}{1.45}
    \begin{tabularx}{\linewidth}{|C{0.35cm}|X|C{0.35cm}|}
    \hline
    \textbf{ID} & \textbf{Text} & \textbf{GT} \\
    \hline
    1 & Yes sure. It was on a Sunday... Sunday evenin...  & 1 \\
    2 & Well, on that day I felt like death, so I did ... & 1 \\
    3 & So yeah, like a couple of months back , thirt...  & 0 \\
    4 & OK. So basically it was months ago, this alre...  & 0 \\
    5 & OK, I sang in Lichfield Cathedral for a weeken... & 1 \\
    \hline
    \end{tabularx}
    }
\end{table}

In our study, the employed dataset was curated from a series of transcripts of statements provided to police prior to court proceedings. The interviews were conducted in the context of analysing verbal veracity cues. Participants were asked to report on a truthful or false event in experiments that took place within the same laboratory located in the United Kingdom and were granted ethical approval by the institutional ethics committee.


The initial dataset underwent a data cleansing process, involving the elimination of irrelevant elements such as fillers (e.g., 'uhm,' 'err'), indicators of participants' non-verbal behaviors (e.g., pauses, smiles), and any contributions from the interviewer. The dataset comprises a total of 687 statements, each accompanied by its corresponding ground truth. The calculated Gini Index of 0.49 indicates a notably balanced distribution of the binary labels. A preview of the data is depicted in Table \ref{tab:dataset}. The last column lists the ground truth, whereby 1 indicates a deceptive statement, and 0 corresponds to the truth.

Comprising a diverse array of statements made by individuals during these preliminary investigative interactions, the dataset aims to capture the varied expressions and complexities inherent in the language used in such interviews. The content encompasses a series of expressions, ranging from straightforward admissions or denials to more subtle explanations, justifications, and appeals. The dataset has been carefully curated to reflect the varied linguistic and argumentative strategies employed in the context of police interviews, mirroring the real-world challenges faced by law enforcement professionals in discerning the credibility of statements.

\subsection{Custom transformer}

To address the task of classifying statements in police interviews, we initially explored the development of a custom transformer model tailored to meet the specific demands of the statement classification domain. The architecture of the model is illustrated in Table \ref{tab:custom_transformer} and incorporates the following elements:

\begin{itemize}
    \item Token and Position Embedding Layer: This layer combines token embeddings and positional embeddings to provide a comprehensive representation of the input sequences.
    \item Transformer Block: The transformer block is employed to capture contextual dependencies within the input data. This block encompasses multi-head self-attention and feed-forward layers, enabling the model to discern the relevance of different tokens and understand complex relationships.
    \item Global Average Pooling: The global average pooling layer is applied to summarise the model's output across the entire sequence, consolidating essential information for downstream processing.
    \item Dropout: Dropout layers are introduced to enhance model robustness and mitigate overfitting.
\end{itemize}

\begin{table}[t]
    \small
    \caption{Custom Transformer Architecture}
    \label{tab:custom_transformer}
    \label{tab:transformer-architecture}
    \renewcommand{\arraystretch}{1.45}
    \begin{tabularx}{\linewidth}{|C{4.1cm}|C{2.1cm}|C{1.35cm}|}
    \hline
    {\textbf{Layer Type}} & \textbf{Output Shape} & \textbf{Param \#} \\
    \hline
    Input & (None, 200) & 0 \\
    Token and Position Embedding & (None, 200, 32) & 223,616 \\
    Transformer Block            & (None, 200, 32) & 10,656 \\
    Global Average Pooling1D     & (None, 32) & 0   \\
    Dropout 2                    & (None, 32) & 0   \\
    Dense 2                      & (None, 16) & 528 \\
    Dropout 3                    & (None, 16) & 0   \\
    Dense 3                      & (None, 1)  & 17  \\
    \hline
    \end{tabularx}
\end{table}
    
The custom transformer model aimed to provide insight to the context of statement classification and to the expected accuracy performance in this task, given the curated dataset. A preliminary evaluation suggested domain-specific advantages, indicating the model's ability to capture subtle language usage pertaining to the data. Nevertheless, the results of this evaluation motivated the decision to explore the possibility of fine-tuning state-of-the-art NLP models, capable of capturing contextual information more effectively and providing enhanced performance and generalisation capabilities. The DistilBERT model was selected for this purpose.

\subsection{DistilBERT Model}

As mentioned above, the transition to DistilBERT was driven by the anticipation of higher performance in light of the significance of accurately classifying statements in the legal context. DistilBERT is a variant of the Bidirectional Encoder Representations from Transformers (BERT) model \cite{devlin2018bert}. It is specifically designed to provide a more computationally efficient but highly effective solution for a wide range of NLP tasks. Its architectural design adheres to the transformer architecture, known for its ability to capture contextual relationships within textual data. DistilBERT inherits the bidirectional context-awareness of BERT while reducing its complexity by distillation techniques, resulting in a more lightweight model, with an approximately 40\% reduced size.

\subsubsection{Applicability}
DistilBERT was selected due to its ability to maintain a competitive performance in NLP tasks, while significantly reducing computational and memory requirements. This efficiency was particularly useful in the case of the curated statement dataset, and facilitated a more accessible model training and deployment process. In addition to its efficiency, the model is pre-trained on a significant amount of text, having a solid foundation on linguistic context and understanding. Finally, the structure of DistilBERT allows straightforward fine-tuning on domain-specific datasets, allowing the model to specialise in this specific task while retaining the benefits of pre-trained knowledge.

\subsubsection{Fine-tuning}

The fine-tuning process was executed with careful consideration of the model configuration and hyperparameter settings. The aim was to adapt the pre-trained DistilBERT model to the specific requirements of statement classification while optimising its performance. The Hugging Face Transformers \cite{wolf2020huggingfaces} library was employed to facilitate the implementation and fine-tuning of the model. The library offers a valuable collection of pre-trained transformer models, fine-tuning pipelines, and utilities for NLP tasks.

\begin{table}[t]
    \small
    \caption {Model Configuration}
    \label{tab:model_config}
    {\renewcommand{\arraystretch}{1.45}
    \begin{tabularx}{\linewidth}{|X|X|}
    \hline
    \textbf{Parameter} & \textbf{Value} \\
    \hline
    Base Model          & distilbert-base-uncased \\
    Activation Function & GELU   \\
    Attention Dropout   & 0.1    \\
    Hidden Dimension (dim) & 768 \\
    Global Dropout Rate    & 0.1 \\
    Fine-Tuning Task       & "sst-2" (Binary Classification) \\
    Hidden Layer Dimension & 3072 \\
    Attention Heads        & 12   \\
    Transformer Layers     & 6    \\
    Dropout                & 0.2  \\
    \hline
    \end{tabularx}
    }
\end{table}

\begin{table}[b]
    \small
    \caption {Training hyperparameters}
    \label{tab:hyperparams}
    {\renewcommand{\arraystretch}{1.45}
    \begin{tabularx}{\linewidth}{|X|X|}
    \hline
    \textbf{Hyperparameter} & \textbf{Value} \\
    \hline
    Loss Function & binary cross-entropy \\
    Optimiser & AdamW \\
    Learning Rate & 0.0002 \\
    Per Device Batch Size & 4 \\
    Weight Decay & 0.01 \\
    Gradient Accumulation Steps & 2 \\
    Epochs & 5 \\
    \hline
    \end{tabularx}
    }
\end{table}

\begin{figure*}[t!]
    \centerline{\includegraphics[width=0.75\linewidth]{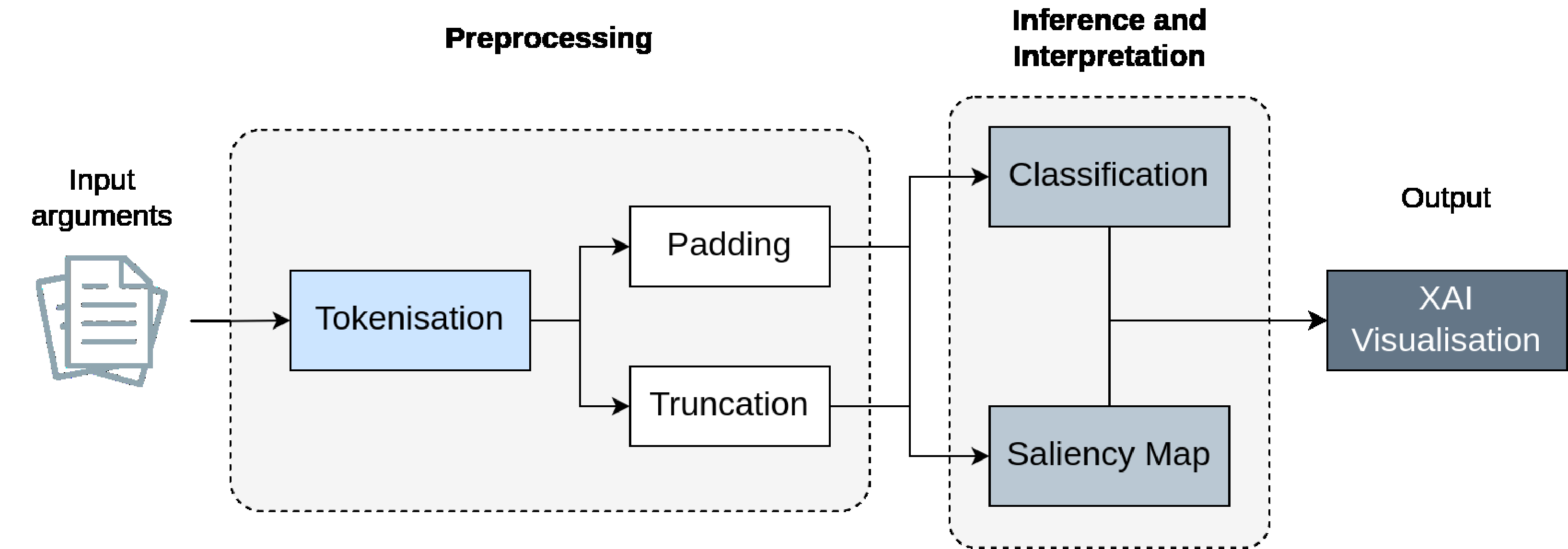}}
    \caption{The end-to-end system pipeline for defensive statement classification and explainability.}
    \label{fig:pipeline}
\end{figure*}

The configuration of the DistilBERT model is outlined in Table \ref{tab:model_config}. It employs the Gaussian error linear units (GELU) activation function, commonly used in Transformers to introduce non-linearity and improve the model's capacity to learn complex patterns \cite{hendrycks2023gaussian}. The GELU function can be approximated as follows:

\begin{equation}
    \text{GELU}(x) = 0.5x (1 + \tanh[\sqrt{\frac{2}{\pi}} \left(x + 0.044715x^3\right)])
\end{equation}

\noindent The GELU function exhibits linear behavior for positive inputs, similar to ReLU activation, but it introduces a smooth, differentiable transition for negative inputs. This transition is achieved by applying a Gaussian distribution to the negative inputs, which helps to prevent the vanishing gradient problem that can occur with ReLU when the input is negative. 

The model was optimised using the binary cross-entropy loss function, which effectively quantifies the divergence between predicted and actual binary labels as follows:

\begin{equation}
    \text{L}(y,p) = -\frac{1}{N} \sum_{i=1}^{N} \left[ y_i \log(p_i) + (1 - y_i) \log(1 - p_i) \right]
\end{equation}

\noindent where $N$ represents the number of samples, $y_i$
is the true label for the $i_{th}$ sample, and $p_i$ is the predicted probability that the $i_{th}$ sample belongs to the positive class.

To promote regularisation during training, the AdamW optimiser was adopted, utilising weight decay to penalise excessive model weights and avoid overfitting. The low selection of per device batch size was made in light of optimising the training process in alignment with the available computational resources, while ensuring the model's efficient convergence and enhancing the overall training performance. The training hyperparameters are outlined in Table \ref{tab:hyperparams}.

The dataset used for training and evaluating the DistilBERT model was divided into training (70\%) and testing (30\%) sets, to assess the model's generalisation performance on unseen data. The training set was used to train the model during the fine-tuning process, while the testing set served as an independent evaluation to determine the model's performance on previously unseen samples and ensure robust insights of the model's effectiveness in real-world scenarios.

\subsection{Explainability through saliency maps}

A crucial aspect of XAI and model interpretability lies in the ability to discern the significance of individual features within the input data. Saliency maps can serve this purpose by quantifying the influence each token in the input sequence has on the model prediction. The central concept in generating a saliency map is determining the sensitivity of the prediction to small perturbations in the input token values. This sensitivity is examined through gradient computations:

\begin{equation}
    \text{Saliency}(\mathbf{t}) = \nabla_{\mathbf{t}} P(y | \mathbf{x}),
\end{equation}

\noindent where $t$ represents a token within the input text $\mathbf{x}$, $P(y | \mathbf{x})$ is the probability distribution over model predictions for the classification task, and $\nabla_{\mathbf{t}}$ denotes the gradient operator with respect to token $t$.

The gradients are captured in a two-step process utilising registered hooks on the model's embeddings. To make the saliency scores more interpretable, we apply L1 normalisation to the computed gradient. This normalisation ensures that the saliency scores sum to 1, enabling direct comparison of the relative importance of each token in the input. The normalised saliency gradient, is calculated as:

\begin{equation}
    \text{Saliency}(\mathbf{t}) = \frac{\nabla_{\mathbf{t}} P(y | \mathbf{x})}{\|\nabla_{\mathbf{t}} P(y | \mathbf{x})\|_1}
\end{equation}

\noindent where $\| \cdot \|_1$ denotes the L1 norm. 

The saliency scores are then mapped to words in the tokenised input text, to generate a saliency map that indicates the most influential words in the input. This mapping provides insights to the model's behaviour and is used in a custom function that highlights the respective words in the text for interpretability. The complete pipeline of our system is illustrated in Figure \ref{fig:pipeline}.

\section{Evaluation}
\label{sec:evaluation}

\subsection{Setup and metrics}

The fine-tuned DistilBERT model was trained and evaluated on an NVIDIA T4 Tensor Core GPU. To comprehensively assess and provide a holistic understanding of the model's capabilities in the context of classifying statements made in police interviews, a range of evaluation metrics were employed, including Accuracy, Precision, Recall, and F1 Score:

\begin{enumerate}

    \item Precision: 
    \begin{equation}
        \text{Precision} = \frac{TP}{TP + FP}
    \end{equation}

    \item Recall: 
    \begin{equation}
        \text{Recall} = \frac{TP}{TP + FN}
    \end{equation}

    \item Accuracy: 
    \begin{equation}
        \text{Accuracy} = \frac{TP + TN}{TP + TN + FP + FN}
    \end{equation}
    
    \noindent where $TP$, $TN$, $FP$, and $FN$ are the true positive, true negative, false positive, and false negative samples respectively.

    \item F1 Score: 
    \begin{equation}
        \text{F1 Score} = \frac{2 \cdot \text{Precision} \cdot \text{Recall}}{\text{Precision} + \text{Recall}}
    \end{equation}
    
\end{enumerate}

To gain further insights to the DistilBERT performance, two additional metrics were used:

\begin{enumerate}

    \item Receiver Operating Characteristic Area Under the Curve (ROC AUC): 

    \begin{equation}
        \text{ROC - AUC} = \int_0^1 TPR(FPR^{-1}(t)) \, dt
    \end{equation}

    \noindent where $TPR$ is the true positive rate and $FPR$ is the false positive rate. The ROC - AUC score is used to quantify the model's ability to distinguish between positive and negative instances.

    \item Average Precision: 

    \begin{equation}
        \text{Average Precision} = \sum_{i=1}^{n} (R_n - R_{n-1}) P_n
    \end{equation}

    \noindent where $R$ is the Recall and $P$ is Precision. Average Precision is used to offer insights into the model's precision-recall trade-offs.

\end{enumerate}
    
\subsection{Results}

\begin{table}[b!]
	\small
    \caption{Classification results}
	\label{tab:results}
    {\renewcommand{\arraystretch}{1.45}
	  \begin{tabularx}{\linewidth}{C{2.57cm}C{1cm}C{1cm}C{1cm}C{1.13cm}}
		\toprule
		Model & Accuracy & Precision & Recall & F1 Score  \\
		\toprule
		Fine-tuned distilbert & 0.8666 & 0.9006 & 0.8265 & 0.8664 \\
		Custom transformer    & 0.8005 & 0.7936	& 0.8126 & 0.8074 \\
		\bottomrule
	  \end{tabularx}
    }
\end{table}

\begin{figure}[t]
    \centerline{\includegraphics[width=0.91\linewidth]{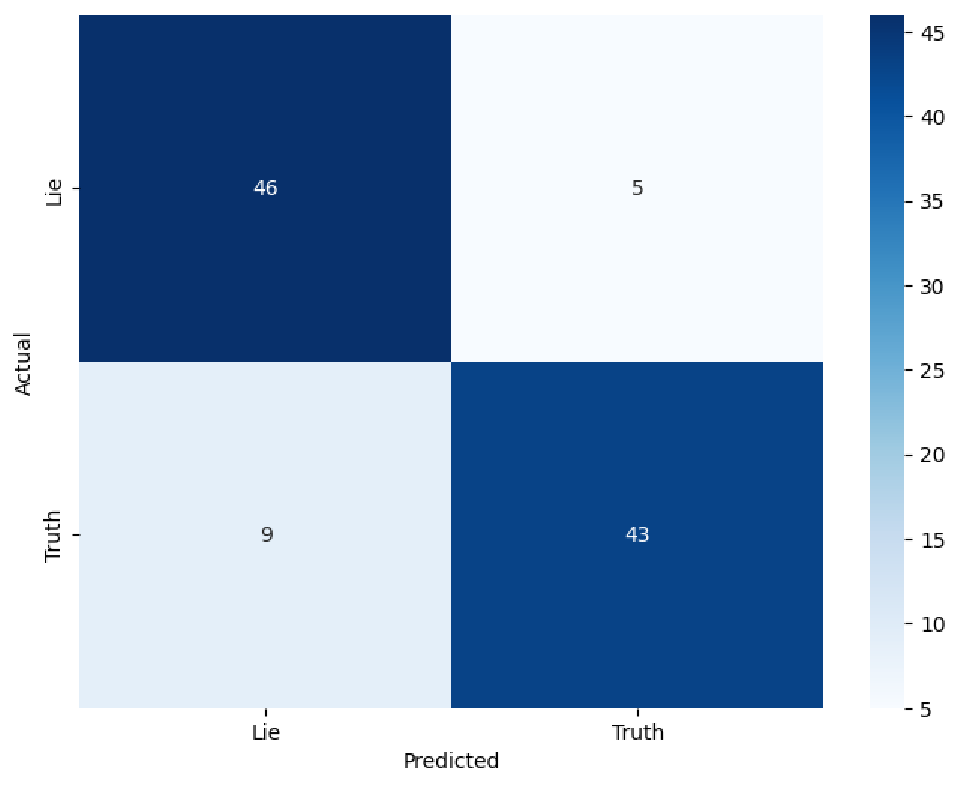}}
    \caption{Confusion matrix of the fine-tuned model.}
    \label{fig:conclusion_matrix}
\end{figure}

The results of the two models across the main metrics are presented in Table \ref{tab:results}. The DistilBERT model achieved a score of 86\% in Accuracy, 90\% in Precision, 82\% in Recall, and 86\% in F1 Score, suggesting the model's robustness and effectiveness in distinguishing between truthful and deceptive claims. The custom transformer's scores of approximately 80\%, while respectable, imply that it may struggle with making accurate predictions in several cases. Figure \ref{fig:conclusion_matrix} presents the confusion matrix on the fine-tuned model.

The DistilBERT model provided high performance as also evidenced by the ROC-AUC score of 0.8671 and the area under the Average Precision of 0.8320. The ROC-AUC score indicates that the model effectively discriminates between positive and negative instances over a range of threshold values in the dataset, while the Average Precision score highlights its ability to correctly classify positive instances while minimising false positives. These results reflect the model's proficiency in capturing the complexity of statements and its potential to make informed and precise predictions in the legal domain.

The results also depict the DistilBERT model's harmonious balance between precision and recall. This is particularly valuable for the legal domain, where both false positives and false negatives can have significant consequences. Ultimately, the results highlight the benefits of leveraging pre-trained transformer models and fine-tuning them for domain-specific tasks, as opposed to utilising custom architectures, thus demonstrating the potential of pre-trained state-of-the-art models in legal defensive statement analysis.

\subsection{Explainable artificial intelligence integration}

\begin{figure}[t!]
    \centerline{\includegraphics[width=0.91\linewidth]{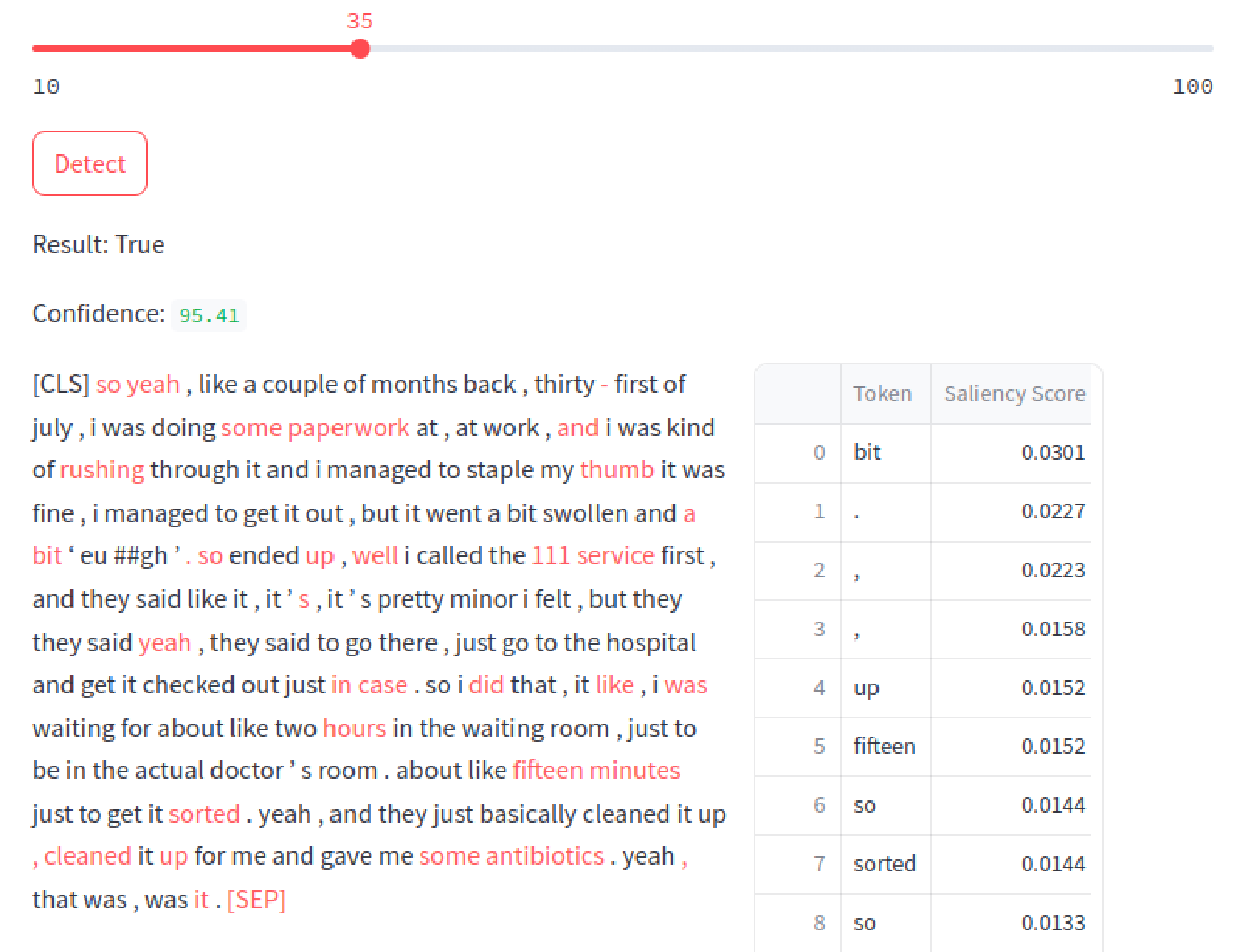}}
    \caption{The XAI interface for defensive statement classification and interpretability.}
    \label{fig:app}
\end{figure}

\begin{figure}[t!]
    \centerline{\includegraphics[width=0.91\linewidth]{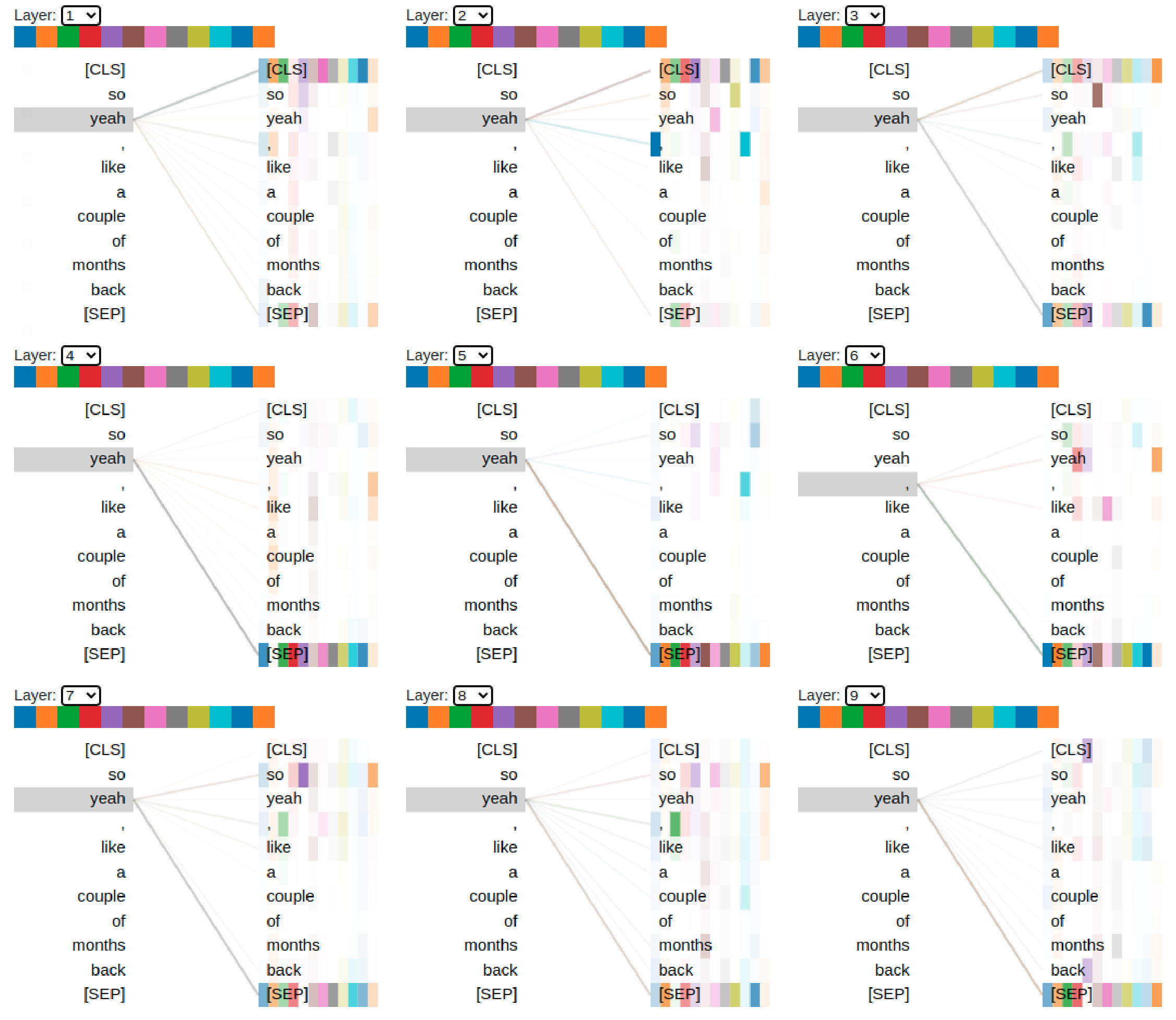}}
    \caption{Visualising the attention in each layer.}
    \label{fig:attention_vis}
\end{figure}

The XAI interface was developed with the primary objective of bridging the gap between theoretical research and its practical application, with a particular emphasis on democratising AI, especially in the legal domain. It provides an accessible and user-friendly platform designed for a diverse audience, including legal practitioners and researchers. Users can submit statements through the interface, and select the number of tokens for interpretability. The deployed model then provides real-time predictions, accompanied by the presentation of the argument with visual emphasis on influential words, as indicated by the saliency map. The XAI interface is presented in Figure \ref{fig:app}. We also provide an example of visualising the attention in each layer in Figure \ref{fig:attention_vis}.

\section{Conclusion}
\label{sec:conclusion}

In the evolving landscape of the legal domain, the integration of NLP methods has become increasingly important, offering innovative solutions to the challenges that legal professionals and researchers encounter. This paper makes significant contributions to the field of argument analysis in statements provided by individuals during interrogations, by introducing an NLP dataset of transcripts from police interviews, allowing for the development and evaluation of our fine-tuned DistilBERT model. The model's high performance in distinguishing truthful from deceptive statements, as well as its capacity to achieve a balance between precision and recall, underscores its potential in making informed and precise predictions in the legal domain. 

Furthermore, the utilisation of state-of-the-art explainability visualisation techniques enhances the model's interpretability, fostering trust and transparency in its decision-making process. The accompanying user-friendly XAI interface further democratises access to our developed tool, bridging the gap between legal practice and research, and making statement analysis more accessible for users of diverse backgrounds.

The experimental results collectively underscore the utility of leveraging pre-trained transformer models and fine-tuning them for domain-specific tasks, as opposed to utilising custom architectures. The success of this approach reinforces the potential of pre-trained state-of-the-art models in the realm of statement analysis in the context of police interrogations, and offers a holistic solution that advances the field, serving as a valuable resource for legal professionals and researchers. 

\bibliographystyle{IEEEtran}

\vspace{0.7cm}
\bibliography{refs}

\end{document}